\documentclass[11pt,a4paper]{article}
\usepackage[hyperref]{emnlp2018}
\usepackage{times}
\usepackage{latexsym}

\usepackage{url}
\usepackage{helvet}  
\usepackage{courier}  
\usepackage{graphicx}  
\usepackage{multirow}
\usepackage{amssymb}
\usepackage{amsmath}
\usepackage{subfigure}
\usepackage{bbm}
\usepackage{array}
\usepackage{bm}
\usepackage{algorithm}  
\usepackage{algorithmic}

\aclfinalcopy 

\title{Attention-Guided Answer Distillation for \\ Machine Reading Comprehension}

\author{
Minghao Hu$^\dag$\thanks{\quad Contribution during internship at Microsoft Research Asia.},
Yuxing Peng$^\dag$,
Furu Wei$^\S$, \\
\bf Zhen Huang$^\dag$, Dongsheng Li$^\dag$, Nan Yang$^\S$, Ming Zhou$^\S$ \\ 
$^\dag$ College of Computer, National University of Defense Technology \\
$^\S$ Microsoft Research Asia  \\
{\tt \{huminghao09,pengyuxing,huangzhen,dsli\}@nudt.edu.cn} \\
{\tt \{fuwei,nanya,mingzhou\}@microsoft.com}
}

\date{}

\begin{document}
\maketitle

\begin{abstract}
Despite that current reading comprehension systems have achieved significant advancements, their promising performances are often obtained at the cost of making an ensemble of numerous models. 
Besides, existing approaches are also vulnerable to adversarial attacks.
This paper tackles these problems by leveraging \emph{knowledge distillation}, which aims to transfer knowledge from an ensemble model to a single model. 
We first demonstrate that vanilla knowledge distillation applied to answer span prediction is effective for reading comprehension systems.
We then propose two novel approaches that not only penalize the prediction on confusing answers but also guide the training with alignment information distilled from the ensemble.
Experiments show that our best student model has only a slight drop of 0.4\% F1 on the SQuAD test set compared to the ensemble teacher, while running 12$\times$ faster during inference.
It even outperforms the teacher on adversarial SQuAD datasets and NarrativeQA benchmark.
\end{abstract}
\section{Introduction}
Machine reading comprehension (MRC), which aims to answer questions about a given passage or document, is a long-term goal of natural language processing.
Recent years have witnessed rapid progress from early cloze-style test~\cite{Hermann15,Hill16} to latest answer extraction test~\cite{Rajpurkar16,joshi2017triviaqa}. 
Several end-to-end neural networks based approaches even outperform the human performance in terms of exact match accuracy on the SQuAD dataset~\cite{Wang17b,wang2018multi,Yu18}.

Despite of the advancements, there are still two problems that impedes the deployment of real-world MRC applications.
First, although effective, current approaches are \emph{not efficient} because superior performances are usually achieved by ensembling multiple trained models. 
For example,~\newcite{Seo17} submit an ensemble model consisting of 12 training runs and~\newcite{Huang17b} boost the result with 31 models. 
The ensemble system, however, has two major drawbacks: the inference time is slow and a huge amount of resource is needed.
Second, existing models are \emph{not robust} since they are vulnerable to adversarial attacks. 
\newcite{Jia17} show that the models are easily fooled by appending an adversarial sentence into the passage.
Such fragility on adversarial examples severely diminishes the practicality of current MRC systems.

One promising direction to address these problems is model compression~\cite{Bucilua06}, which attempts to compress the ensemble model into a single model that has comparable performances. 
Particularly, the \emph{knowledge distillation} approach~\cite{Hinton14} has been proposed to train a student model with the supervision of a teacher model.
Such idea is further explored to enhance the generalizability and robustness of image recognition systems~\cite{Papernot16}.
Some subsequent works attempt to transfer teacher's intermediate representation, such as feature map~\cite{Romero15,Zagoruyko17}, neuron selectivity~\cite{Huang17} and so on, to provide additional supervisions.

\begin{figure*}
  \centering
  \subfigure[Gold answers (\textcolor{red}{red}) versus confusing answers (\textcolor{blue}{blue}) in the SQuAD and adversarial SQuAD datasets.]{
    \label{fig1:a} 
    \includegraphics[width=3.1in]{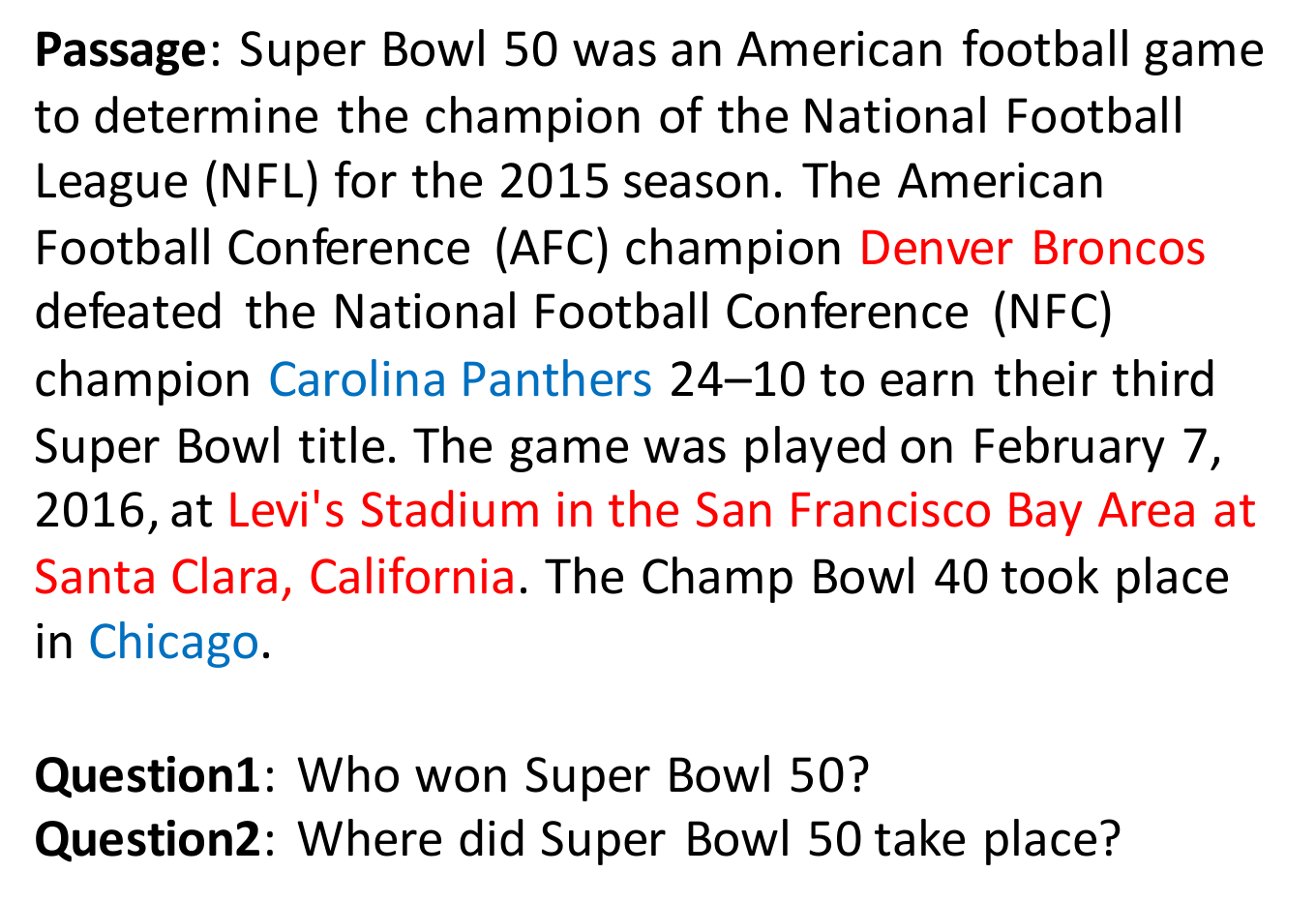}}
  \hspace{0.1in}  
  \subfigure[The knowledge biased towards the confusing answer is distilled from the teacher to the student.]{
    \label{fig1:b} 
    \includegraphics[width=2.8in]{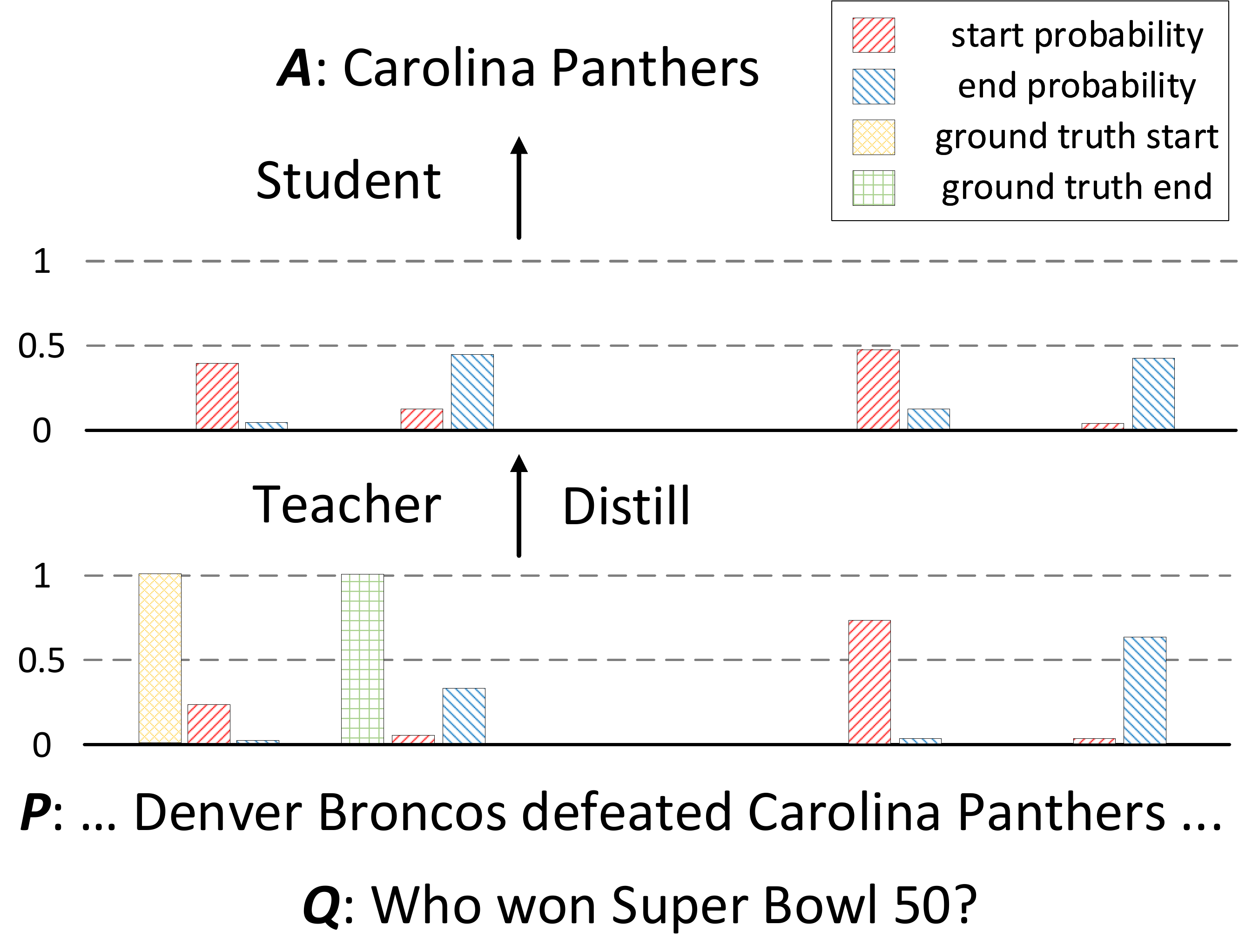}}
  \vspace{-0.2cm}
  \caption{An illustration of confusing answer and biased distillation in machine reading comprehension.}
  \label{fig1} 
  \vspace{-0.2cm}
\end{figure*}

In this paper, we present the first work to investigate knowledge distillation in the context of MRC, to improve efficiency and robustness simultaneously.
We first apply the standard knowledge disillation to MRC models, by mimicking output distributions of answer boundaries from an ensemble model, and observe consistent improvements upon a strong baseline.
We then propose two novel distillation approaches to further transfer knowledge between the teacher and the student.

First, we introduce \emph{answer distillation}, which penalizes the most confusing answer with a margin loss, to deal with the problem that biased knowledge misleads the student into incorrect predictions.
We find that in MRC datasets there exists many confusing answers (Figure \ref{fig1:a}), which match the category of true answers but are semantically-contradicted to the question, and an extreme case is the adversarial example.
Once the teacher produces biased probabilities towards these distractors, inaccurate distributions will be distilled and later used to supervise the student.
As a result, the student could produce over-confident wrong predictions.
We refer to this problem as \emph{biased distillation}, and Figure \ref{fig1:b} gives an example.
To address this problem, our approach distilles the boundary of the strongest distractor from the teacher, and explicitly informs the student to decrease its confidence score. 
This forces the student to produce comparable and unbiased distributions between candidate answers.

Second, we present \emph{attention distillation} that aims to match the attention distribution between the teacher and the student. 
We notice that neural attention plays an important role in MRC tasks by enabling the model to capture complex interactions between the question and the passage. 
Compared to other forms of intermediate representation such as feature map and neuron selectivity, attention distribution is more compressed and more informative in reflecting semantic similarities of input text pairs. 
Hence, by mimicking the word alignments from an ensemble model, we expect that the student can learn to attend more precisely with better compression efficiencies.

We evaluate our approach on the Stanford Question Answering Dataset (SQuAD)~\cite{Rajpurkar16}, the adversarial SQuAD dataset~\cite{Jia17} and the NarrativeQA benchmark~\cite{Kovcisky2017}.
Compared to the ensemble teacher, the single student model trained with our approach has a slight drop of 0.4\% F1 on the SQuAD test set, while running 12$\times$ faster during inference.
The student even outperforms the teacher on the adversarial SQuAD dataset, and surpasses the teacher in terms of Bleu-1 score on the NarrativeQA benchmark.
\section{Background}
\begin{figure*}
\begin{center}
\includegraphics[width=6.3in]{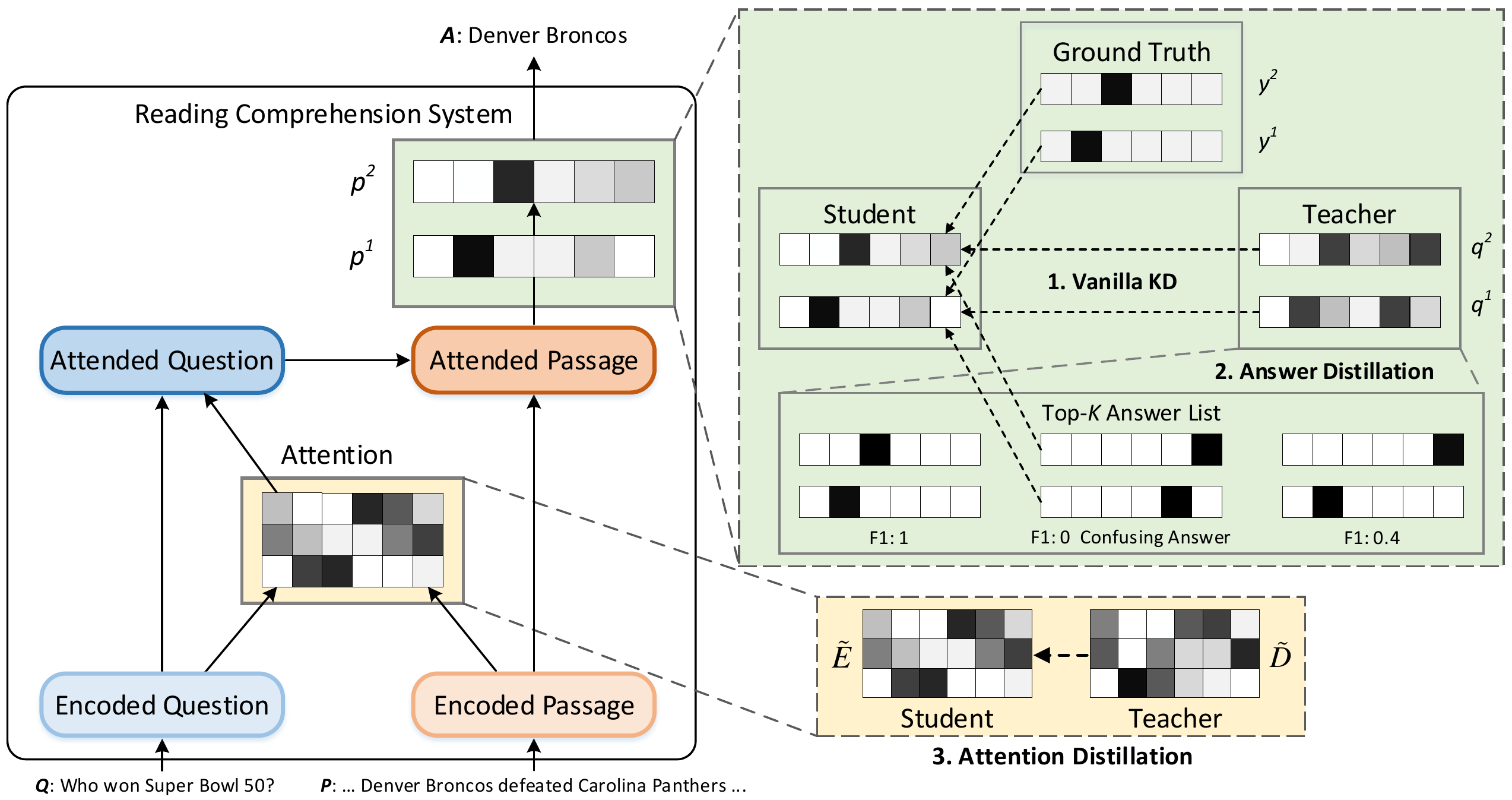}
\end{center}
\vspace{-0.2cm}
\caption{Overview of our approaches. 
In vanilla knowledge distillation (\textcolor[RGB]{84,130,53}{green}), the cross entropy is minimized between the student/teacher distributions of answer positions. 
In answer distillation (\textcolor[RGB]{84,130,53}{green}), the student is trained to penalize the most confusing answer distilled from the teacher.
In attention distillation (\textcolor[RGB]{255,217,102}{yellow}), mean-squared error is minimized between the student/teacher attention distributions.
Darker color denotes higher probability.}
\vspace{-0.2cm}
\label{fig2}
\end{figure*}

\subsection{Machine Reading Comprehension}	\label{sec:1}
In the extractive MRC task, the question and the passage are described as sequences of word tokens, denoted as $Q=\{q_i\}_{i=1}^n$ and $P=\{p_j\}_{j=1}^m$ respectively. 
The task is to predict an answer $A$, which is constrained as a segment of text in the passage: $A=\{p_j\}_{j=k}^l$. 

To model complicated interactions between the question and the passage, the attention mechanism~\cite{Bahdanau15} has been widely used. 
Let $V=\{v_i\}_{i=1}^n$ and $U=\{u_j\}_{j=1}^m$ denote the encoded question/passage representations, where $v_i$ and $u_j$ are both dense vectors. 
A similarity matrix $E \in\mathbb{R}^{n \times m}$ can be computed as:
\begin{eqnarray}	
	E_{ij} = f(v_i, u_j) \nonumber
\end{eqnarray}
where $f$ is a scalar function producing the similarity between two inputs. 
Let $\mathrm{softmax}(x)$ denote the softmax function that normalizes the vector $x$. 
Then the attention distribution across the question for the $j$-th passage word is computed as $\tilde{E}_j = \mathrm{softmax}(E_{:j})$. 
This attentive information is commonly used to summarize the entire question as a single vector that is later fused back into the $j$-th passage word~\cite{Seo17}, resulting in an attended passage representation.

Finally, based on the attended passage, previous works usually apply a pointer network~\cite{Vinyals15} to produce two probabilities $p^1$ and $p^2$ that indicate the answer start and end positions:
\begin{eqnarray} 
	p(A|Q, P) = p^1(k|Q, P) p^2(l|k,Q,P) \nonumber
\end{eqnarray}
where $p$ is the model distribution.

\subsection{Knowledge Distillation} \label{sec:2}
The knowledge distillation framework~\cite{Hinton14} consists of two networks: a \emph{teacher} $T$, which is a pre-trained large model or an ensemble of multiple models, and a \emph{student} $S$, which is a smaller network that learns from the teacher. The idea is to supervise the student with not only ground truth labels but also output distributions of the teacher.
Concretely, given a data set of examples of the form $(X, Y)$, the cross-entropy loss can be minimized to learn a multi-class classifier:
\begin{gather}	
	\mathcal{L}_{CE}(\theta) = - \sum_{k=1}^m Y_k \log p(k|X;\theta) \nonumber
\end{gather}
where $p$ is the model distribution that is parameterized by $\theta$, and $m$ indicates the number of classes. 
The standard knowledge distillation is to replace ground truth $Y$ with a soft probability distribution $q$ generated by the teacher as: 
\begin{gather}
	q=\mathrm{softmax}(\frac{\alpha}{\tau}), \quad p=\mathrm{softmax}(\frac{\beta}{\tau}) \nonumber \\
	\mathcal{L}_{KD}(\theta_S) = - \sum_{k=1}^m q(k|X;\theta_T) \log p(k|X;\theta_S) \nonumber
\end{gather}
where $\alpha$ and $\beta$ are pre-softmax logits for teacher and student respectively, and $\tau$ is a temperature coefficient that is normally set to 1. A higher $\tau$ produces a softer probability distribution, and thus, provides more information about the relative similarity between classes.
As Hinton et al. \shortcite{Hinton14} suggested, the above losses can be jointly optimized as follows:
\begin{eqnarray}	
	\mathcal{L}(\theta_S) = \mathcal{L}_{CE}(\theta_S) + \lambda \mathcal{L}_{KD}(\theta_S) \nonumber
\end{eqnarray}
where $\lambda$ is usually set as $\tau^2$ since the magnitudes of gradients produced by $\mathcal{L}_{KD}$ scale as $1 / \tau^2$.
\section{Attention-guided Answer Distillation}

Figure \ref{fig2} gives an overview of our distillation framework, which mainly consists of three approaches including vanilla knowledge distillation, answer distillation and attention distillation.
First, we explore standard knowledge distillation in the context of MRC by replacing one-hot labels with soft output distributions extracted from the teacher. 
We then distill the span of the most confusing answer from the teacher, so that the student can learn to distinguish gold answers from the distractor.
Next, we utilize teacher's attention distributions to guide the student's training process for forcing the student to attend more precisely.
Finally, the student is jointly trained with the above distilled knowledge.

\subsection{Vanilla Knowledge Distillation}
The standard training method for MRC models is to minimize the cross entropies on two answer positions~\cite{Wang17a}:
\begin{eqnarray}
	\mathcal{L}_{CE} = -\sum_{k=1}^m \sum_{l=1}^m y_k^1 \log p^1(k) + y_l^2 \log p^2(l|k) \nonumber
\end{eqnarray}
where $y^1$ and $y^2$ are one-hot labels for the answer start and end positions respectively, and $m$ refers to the passage length. We denote $p^1(k|Q, P)$ as $p^1(k)$ and $p^2(l|k,Q, P)$ as $p^2(l|k)$ for abbreviation. 

Following the standard procedure of knowledge distillation in Section \ref{sec:2}, we can replace one-hot labels with output distributions of answer start and end positions predicted by the teacher as:
\begin{align}
	\mathcal{L}_{KD} = - \sum_{k=1}^m \sum_{l=1}^m & q^1(k)  \log p^1(k) + \nonumber \\
	& q^2(l|k) \log p^2(l|k)  \nonumber
\end{align}
Here, we first consider the teacher as a single model, and later extend it to the ensemble scenario in Section \ref{sec:3}.

\subsection{Answer Distillation}
Vanilla knowledge distillation allows the student to learn relative similarities between answer candidates at position level. 
However, the student can suffer from the biased distillation problem once the teacher makes wrong predictions towards confusing answers. 
As a result, the model may be over-confident on the distractors during inference. 
Therefore, it is necessary to produce confidence scores that are comparable and unbiased between candidate answers.
Since there usually exists one most confusing candidate answer, we hence consider explicitly informing the student about its boundary so as to relatively decrease the corresponding confidence.

Specifically, we perform inference with the teacher network to get a top-$K$ answer list $\mathcal{A_K}$ for each example, and measure the word overlap between the ground truth $A^*$ and each answer candidate $A_i$ from the list. 
The one that shares no overlap with gold answers and has the highest confidence score is chosen as the \emph{confusing answer}.
We use the F1 scoring function $F_1(A^*, A_i)$ as the measurement of word overlap and use the probability $q(A_i)$ as the confidence score.
If the F1 scores of all top-$K$ answers are larger than 0, then we argue that the teacher makes a good prediction and therefore do not distill the confusing answer for this example.
The above process is performed on the entire training set to annotate all potential confusing answers.

Once the process is finished, we force the student to produce confidence scores of gold answers that are distinguishable from the score of confusing answer, by minimizing a margin ranking loss~\cite{bai2010learning} as:
\begin{align}
	\mathcal{L}_{ANS} = & \max (0, 1 - \beta_k^1 + \beta_i^1) + \nonumber \\ 
	&\max (0, 1 - \beta_l^2 + \beta_j^2) \nonumber
\end{align}
where $\beta$ is the student's pre-softmax logit, $k$ and $l$ indicate the boundary of gold span while $i$ and $j$ refer to the confusing boundary. 
With this loss, we penalize the strongest distractor on which the model is over-confident, and encourage the true answer that is underestimated by the student.

\subsection{Attention Distillation}
The above distillation approaches allow knowledge to be transferred through teacher's outputs. 
However, we would like to distill not only the final outputs but also some intermediate representations~\cite{Romero15}, in order to provide additional supervised signals for training the student network. 
The standard approach is to regress the student's intermediate passage representation to the teacher's corresponding representation as a pre-training step.
Nevertheless, this approach is cumbersome in that the dimension of passage representation, denoted as $h \times m$ where $h$ is the hidden size, is quite large, leading to a huge amount of resource consumption as we need to distill all training examples.
Since neural attention is more compressed and contains rich informantion about where to attend, we hence propose to match the attention distribution between  the teacher and the student instead. 

Following the notation in Section \ref{sec:1}, let $\tilde{D}_j$ and $\tilde{E}_j$ denote the attention distribution across the question for the $j$-th passage word in teacher and student repectively. 
We define a mean-squared loss as:
\begin{eqnarray}
	\mathcal{L}_{ATT} = \frac{1}{2} \sum_{j=1}^m ||\tilde{D}_j - \tilde{E}_j||^2  \nonumber
\end{eqnarray}
Compared to regressing the hidden representation, attention distillation brings in two benefits: 
1) the dimension of similarity matrix is $n \times m$, where the question length $n$ is significantly smaller than the hidden size $h$, resulting in a better compression efficiency; 
2) the student can learn to attend more precisely as neural attention directly reflects semantic similarities between input text pairs.

\subsection{Joint Training} \label{sec:3}
So far, we define three losses to transfer knowledge between MRC models. Next we consider using a joint objective to train the student network as:
\begin{eqnarray} \label{eq:1}
	\mathcal{L} = \mathcal{L}_{CE} + \lambda \mathcal{L}_{KD} + \gamma \mathcal{L}_{ANS} + \delta \mathcal{L}_{ATT}
\end{eqnarray}
where $\lambda$ is set as $\tau^2$ to scale gradients, $\gamma$ and $\delta$ are two hyper-parameters that control the task-specific weights.

In addition, since the knowledge is distilled from an ensemble model, we need to integrate multiple outputs into an unified label. For vanilla knowledge distillation and attention distillation, we use an arithmetic mean of their individual predicted distributions as the final soft label. As for answer distillation, we choose the confusing answer that has the highest confidence score among all models.

In summary, the entire training procedure is:
1) train an ensemble teacher model; 
2) distill knowledge using the teacher on the training set;
3) integrate multiple types of knowledge to build a new training set;
4) train a single student model with Equation \ref{eq:1} on the new dataset.

\section{Experiments}

\subsection{Datasets and Metrics}
We conduct experiments on the following three datasets.

\noindent{\bf SQuAD} (Stanford Question Answering Dataset)~\cite{Rajpurkar16} is a machine comprehension dataset containing $100,000$+ questions that are annotated by crowd workers on $536$ Wikipedia articles.
The answer to each question is always a span in the corresponding passage. 

\noindent{\bf Adversarial SQuAD}~\cite{Jia17} is a dataset aiming to test whether the model truely understands the text by appending an adversarial sentence to the passage. 
A strong confusing answer is constructed in the adversarial sentence to distract the answer prediction.

\noindent{\bf NarrativeQA}~\cite{Kovcisky2017} is a benchmark proposed for story-based reading comprehension. 
The answers in this dataset are handwritten by human annotators based on a short summary. 
Following~\newcite{Tay2018}, we compete on the summary setting to compare with reported baselines. We choose the span that achieves the highest Rouge-L score with respect to the gold answers as labels for training. 

We use the official metrics to perform the evaluation. 
Specifically, for both SQuAD and adversarial SQuAD datasets, we report exact match (EM) and F1 scores. 
As for the NarrativeQA benchmark, the metrics are Blue-1, Bleu-4 and Rouge-L.

\subsection{Implementation Details}
We implement the Reinforced Mnemonic Reader (RMR)~\cite{Hu17} as our base model, which contains the standard attention mechanism and the two-step answer prediction described in Section \ref{sec:1}.
Therefore, our distillation approaches can be seamlessly used, and the improvement of our approaches can be directly compared to the baseline.
We use slightly different network architectures and hyper-parameters for different datasets.
More specifically, we use the original configuration for SQuAD and adversarial SQuAD datasets.
As for the NarrativeQA benchmark, we truncate the passage to the first $800$ words and use a batch size of $32$. 
Besides, we also remove ELMo embeddings~\cite{Elmo18} and reduce the number of aligning layer to $2$ to avoid out-of-memory problem.

\begin{table}
\begin{center}
\begin{tabular}{lcccc}
\hline 
\multirow{2}*{ Model } & \multicolumn{2}{c}{ Dev } & \multicolumn{2}{c}{ Test } \\
 & EM & F1 & EM & F1 \\ 
\hline
LR Baseline$^1$  & 40.0 & 51.0 & 40.4 & 51.0 \\ 
FusionNet$^2$  & 75.3 & 83.6 & 76.0 & 83.9 \\
BiSAE$^3$  & 77.9 & 85.6 & 78.6 & 85.8 \\
R-Net+$^4$ & - & - & 79.9 & 86.5 \\
SLQA+$^5$  & 80.0 & 87.0 & 80.4 & 87.0 \\
QANet$^6$  & - & - & 80.9 & 87.8 \\
\hline
BiSAE (E)  & 79.6 & 86.6 & 81.0 & 87.4 \\
R-Net+ (E) & - & - & 82.6 & 88.5 \\
SLQA+ (E)  & 82.0 & 88.4 & 82.4 & 88.6 \\
QANet (E)  & - & - & 82.7 & 89.0 \\
\hline
RMR$^7$   & 78.9 & 86.3 & 79.5 & 86.6 \\
RMR (E)   & \bf 81.2 & \bf 87.9 & \bf 82.3 & \bf 88.5 \\
RMR + A2D & 80.3 & 87.5 & 81.5 & 88.1 \\
\hline
\end{tabular}
\caption{\label{table1} Comparison of different approaches on the SQuAD test set, extracted on May 9, 2018:~\newcite{Rajpurkar16}$^1$,~\newcite{Huang17b}$^2$,~\newcite{Elmo18}$^3$,~\newcite{Wang17b}$^4$,~\newcite{wang2018multi}$^5$,~\newcite{Yu18}$^6$ and~\newcite{Hu17}$^7$. BiSAE refers to BiDAF + Self Attention + ELMo. (E: ensemble model)}
\vspace{-0.4cm}
\end{center}
\end{table}

We run $12$ single models with the identical architecture but different initial parameters to obtain the ensemble model.
The student has the same network architecture as the teacher. 
We reuse the models trained on the SQuAD dataset to test on adversarial SQuAD datasets, but we retrain another pair of teacher and student models for the NarrativeQA benchmark.
The temperature $\tau$ is tuned among $[1, 2, 3, 5]$ and is set as $2$ by default. 
The weight $\gamma$ is set to $0.3$ and $\delta$ is $0.1$. 
$K$ is set to $4$ for generating the top-$K$ answer list. 
All experiments and runtime benchmarks are tested on a single Nvidia Tesla P100 GPU. 
Our approach is denoted as A2D for abbreviation.

\subsection{Main Results}
In this section we report main results on three MRC datasets\footnote{\href{https://worksheets.codalab.org/worksheets/0xfa2548e70c2540ca8c2d26d7dd402f7f/}{https://worksheets.codalab.org/worksheets/ 0xfa2548e70c2540ca8c2d26d7dd402f7f/}.}. 
Since our main purpose is to compress the ensemble model into a single model that possesses better efficiency and robustness, we focus on comparing our approach with the ensemble baseline.

\begin{table}
\begin{center}
\begin{tabular}{lcccc}
\hline 
\multirow{2}*{ Model } & \multicolumn{2}{c}{ AddSent } & \multicolumn{2}{c}{ AddOneSent } \\
 & EM & F1 & EM & F1 \\ 
\hline
LR Baseline & 17.0 & 23.2 & 22.3 & 30.4 \\
BiSAE & 38.7 & 44.4 & 48.0 & 54.7 \\
SLQA+ & - & 52.1 & - & 62.7 \\
DCN+ (MINI)$^1$ & 52.2 & 59.7 & 60.1 & 67.5 \\
\hline
FusionNet (E) & 46.2 & 51.4 & 54.7 & 60.7 \\
SLQA+ (E) & - & 54.8 & - & 64.2 \\
\hline
RMR & 53.0 & 58.5 & 60.9 & 67.0 \\
RMR (E) & 56.0 & 61.1 &  62.7 &  68.5 \\
RMR + A2D & \bf 56.0 & \bf 61.3 & \bf 63.3 & \bf 69.3 \\
\hline 
\end{tabular}
\caption{\label{table2} Comparison of different approaches on two adversarial SQuAD datasets:~\newcite{min2018efficient}$^1$. (E: ensemble model)}
\vspace{-0.4cm}
\end{center}
\end{table}

We present the results on the test set of SQuAD in Table \ref{table1}.
We can see that the student network is able to compete with the ensemble model with only a slight drop of 0.4\% on F1. 
Moreover, nearly 80\% of the improvement in terms of F1 achieved by the ensemble model is successfully transferred to the distilled model, which indicates the effectiveness of our approach.

To validate the effect of our approach on enhancing robustness, we show the results on two adversarial SQuAD datasets, namely AddOneSent and AddSent, in Table \ref{table2}.  
As we can see, the improvement on adversarial data is much higher than the one on original SQuAD dataset: the student network successfully surpasses the teacher.
The significant improvement comes from the fact that there exists much more confusing answers in adversarial datasets, and hence the baseline is more likely to be over-confident on distractors.
Our approach, however, explicitly decreases distractor's confidence, thus yielding more robust predictions against adversarial examples.

\begin{table}
\begin{center}
\begin{tabular}{lccc}
\hline 
Model & Bleu-1 & Bleu-4 & Rouge-L \\
\hline
Seq2Seq & 15.9 & 1.3 & 13.2 \\ 
AS Reader$^{1}$ & 23.2 & 6.4 & 22.3 \\
BiDAF$^{2}$ & 33.7 & 15.5 & 36.3\\
BiAttention$^{3}$ & 36.6 & 19.8 & 41.4\\
\hline
RMR & 48.4 & 24.6 & 51.5\\
RMR (E) & 50.1 & \bf 27.5 & \bf 53.9 \\
RMR + A2D & \bf 50.4 & 26.5 & 53.3  \\
\hline 
\end{tabular}
\caption{\label{table3} Comparison of different approaches on the NarrativeQA test set using summaries:~\newcite{Kadlec2016}$^1$,~\newcite{Seo17}$^2$ and~\newcite{Tay2018}$^3$. (E: ensemble model)}
\vspace{-0.4cm}
\end{center}
\end{table}

To verify the generalizability of our approach among various datasets, we further detail the results on the test set of NarrativeQA in Table \ref{table3}, 
Despite that both of our single baseline and ensemble model have already achieved top results, the student's performance can still be boosted using our approach, even outperforming the teacher in terms of Bleu-1 score.

\subsection{Speedup over Ensemble Model}
Next, in order to show the improvement on efficiency brought by our approach, we compare the inference speedup of the student against the teacher in Table \ref{table4}.
Since we use 12 single models to make the ensemble, we can easily see that the student is nearly 12$\times$ faster than the teacher. 
Besides, compared to the baseline, the student has nearly the same number of parameters, demonstrating that our approach does not introduce additional computation complexity.

\begin{table}
\begin{center}
\begin{tabular}{lcccc}
\hline 
 & Params & Time & Speedup \\
\hline
RMR (E)     & 6.9m$\times$12 & 118.2 & - \\
RMR + A2D   & 6.9m & \bf 9.6 & \bf 12.3\bm{$\times$} \\
\hline
\end{tabular}
\caption{\label{table4} Comparison between the ensemble teacher and the single student on the SQuAD dev set. Time denotes number of minutes needed to perform the entire inference. (E: ensemble model)}
\vspace{-0.4cm}
\end{center}
\end{table}

\subsection{Ablation Study}
To get better insights of our distillation approaches, we conduct in-depth ablation study on both the development set of SQuAD and the AddSent set of adversarial SQuAD. 
We mainly focus on the F1 score since it is used as the main metric on the SQuAD leaderboard.

Table \ref{table5} shows the ablation results. 
First, we evaluate the vanilla knowledge distillation by removing the KD loss. 
We find that this loss contributes a lot on both datasets, implying that standard knowledge distillation is helpful for MRC models. 
Next we ablate the attention loss (ATT), and observe a smaller performance degradation on both datasets, which suggests that matching attention distributions is also beneficial but less effective.
We then test the effect of removing answer distillation (ANS), and discover that although the impact on the SQuAD dev set is relatively small, the influence on the AddSent dataset is quite large. We argue that this is because the model trained with answer distillation can better handle the biased distillation problem, which is more severe in the adversarial dataset.
Finally, we replace the joint training process with a stage-wise fashion proposed by~\newcite{Romero15}. Concretely, we first warm up the student by matching the attention distribution as a pre-training step, and then minimize the rest of losses to train the model. The result, however, shows that this strategy does harm to the performance. We think the reason may be that the pre-training step leads the model into a local minima.

\begin{table}
\begin{center}
\begin{tabular}{lcccc}
\hline 
\multirow{2}*{  } & \multicolumn{2}{c}{ Dev } & \multicolumn{2}{c}{ AddSent } \\
 & F1 & $\Delta$F1 & F1 & $\Delta$F1 \\ 
\hline
RMR + A2D   & 87.5 & - & 61.3 & - \\
\ \ - Vanilla KD  & \underline{86.8} & \underline{-0.7} & 60.1 & -1.2 \\
\ \ - ATT & 87.0 & -0.5 & 60.4 & -0.9 \\
\ \ - ANS & 87.1 & -0.4 & \underline{59.8} & \underline{-1.5} \\
\ \ - Joint Training & 87.3 & -0.2 & 60.8 & -0.5 \\
\hline
\end{tabular}
\caption{\label{table5} Ablation study of different knowledge distillation approaches on SQuAD dev set and AddSent set.}
\vspace{-0.4cm}
\end{center}

\end{table}

\begin{figure*}
  \centering
  \subfigure[Alignments before/after distillation. Using A2D, the attention distributions concentrate less on superfical clues around the confusing answer ``\emph{56.2\%}'' (e.g., ``\emph{population}'' to ``\emph{inhabitants}'', ``\emph{Protestant}'' to ``\emph{Catholics}'' and so on.).]{
    \label{fig3:a} 
    \includegraphics[width=6.2in]{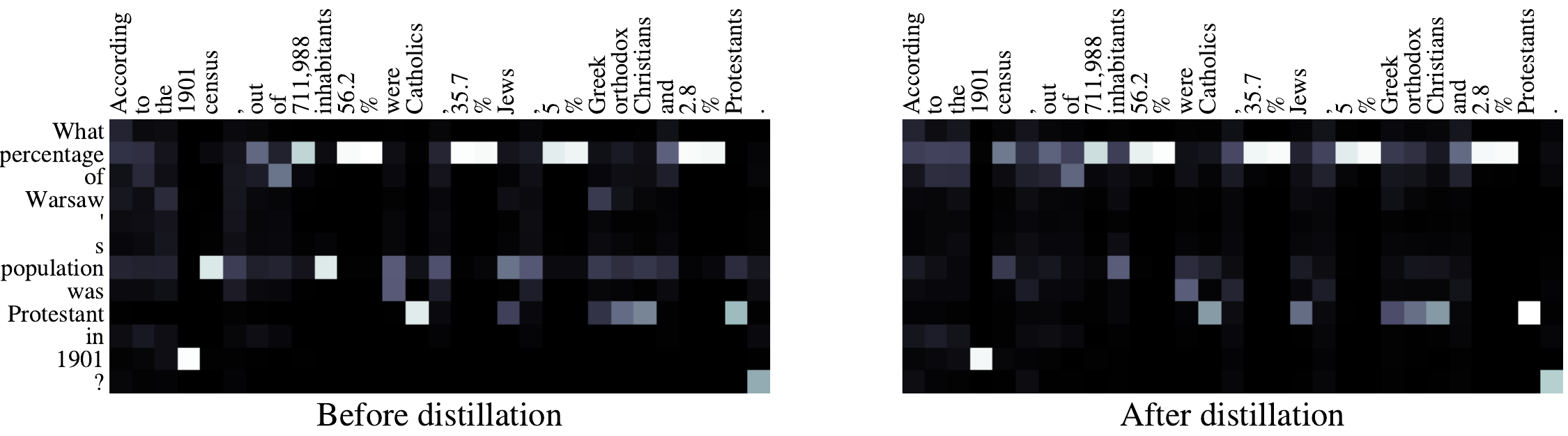}}
  \hspace{0.3in}  
  \subfigure[The base model points to the confusing answer (``\emph{56.2\%}''), while the distilled model predicts the correct one (``\emph{2.8\%}'').]{
    \label{fig3:b} 
    \includegraphics[width=6.2in]{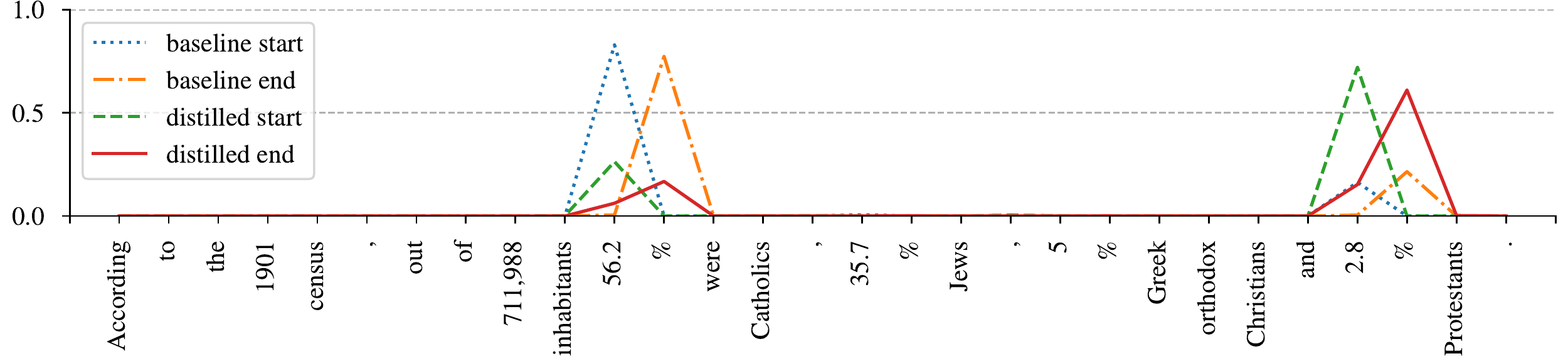}}
  \vspace{-0.2cm}
  \caption{A case study between the base model and the distilled model.}
  \label{fig3} 
  \vspace{-0.2cm}
\end{figure*}

\subsection{Analysis and Discussion}
We now give a qualitative analysis on our approach by visualizing attention distributions and output probabilities for both of the baseline and the distilled model.
From Figure \ref{fig3:a} we can see that, both models are good at finding candidate answers, such as ``\emph{56.2\%}'' and ``\emph{2.8\%}'', according to the key question word ``percentage''. 
Nevertheless, the base model pays more attention on question-passage word pairs around the confusing answer ``\emph{56.2\%}''. 
As a result, the model is more likely to produce a high confidence score on ``\emph{56.2\%}'', as shown in Figure \ref{fig3:b}.
Using our approach, however, leads to less concentrations on superfical clues around the confusing answer.
Instead, the distilled model is able to focus more on the critical alignments (e.g., ``\emph{Protestant}'' to ``\emph{Protestants}''), and therefore predicts the correct answer ``\emph{2.8\%}''.

\begin{figure*}
  \centering
  \subfigure[Results on the SQuAD dev set.]{
    \label{fig4:a} 
    \includegraphics[width=2.9in]{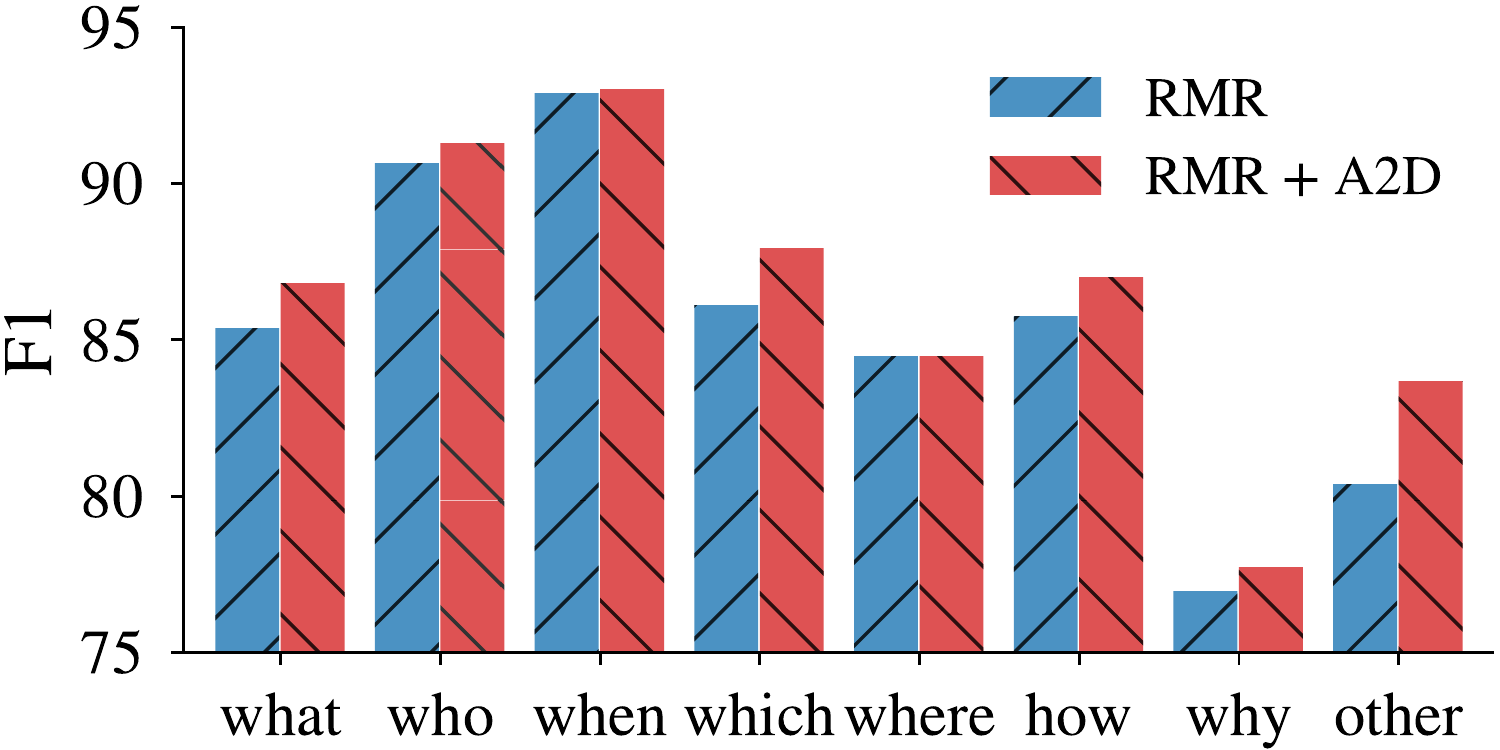}}
  \hspace{0.3in}  
  \subfigure[Results on the AddSent set.]{
    \label{fig4:b} 
    \includegraphics[width=2.9in]{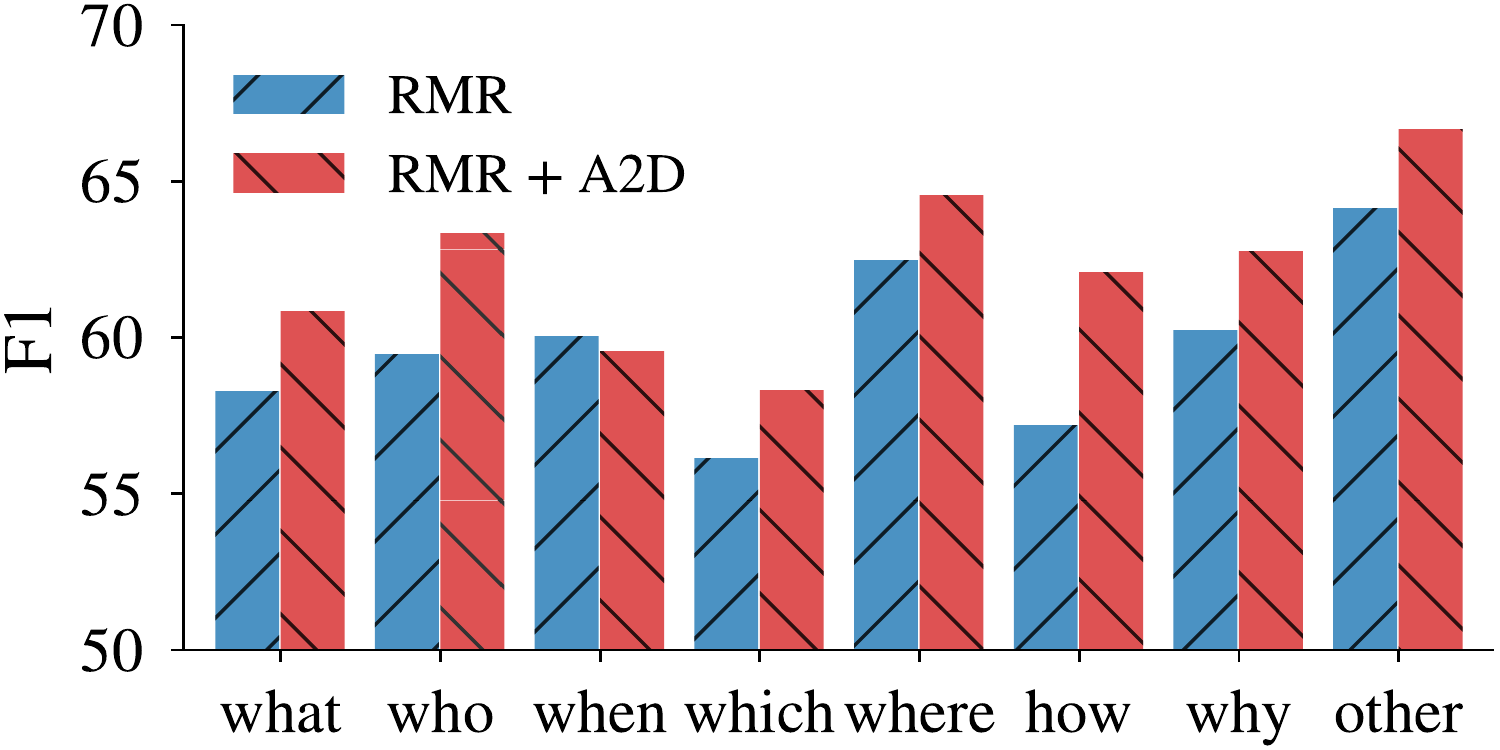}}
  \vspace{-0.2cm}
  \caption{Comparison of different question types between the base model and the distilled model.}
  \label{fig4} 
  \vspace{-0.2cm}
\end{figure*}

To show on which type of questions our approach is doing better, we report the results in Figure \ref{fig4}. 
We can see that our approach yields consistent performance gain over the baseline across different question types on various datasets.
Particularly, A2D provides an obvious advantage for question types such as ``\emph{what}'', ``\emph{who}'', ``\emph{which}'', ``\emph{how}'', ``\emph{why}'' and so on.

\section{Related Work}
\noindent{\bf Machine Reading Comprehension.} 
Benefiting from large-scale machine reading comprehension (MRC) datasets~\cite{Hermann15,Hill16,Rajpurkar16,joshi2017triviaqa}, end-to-end neural networks have achieved promising results~\cite{wang2018multi,Yu18}.
\newcite{Wang17a} combine the match-LSTM with pointer networks to predict the answer boundary.
\newcite{Wang17b} match the context aginst itself to refine the passage representation.
Later, a variety of attention mechanisms have been proposed, such as bi-attention~\cite{Seo17}, coattention~\cite{Xiong17}, fully-aware attention~\cite{Huang17b} and reattention~\cite{Hu17}.
Among these works, two common traits can be summarized as: 1) compute a similary matrix between the question and the passage; 2) sequentially predict the answer start and end positions.
Our proposed approach is a simple and effective adaptation to existing models by taking advantage of these traits, and do not complicate previous works more than necessary.

\noindent{\bf Efficiency and Robustness in MRC.} 
Improving efficiency and robustness for reading comprehension system has attracted a lot of interest in recent years.
For efficiency, previous works mostly concentrate on how to scale passage-level models to large corpora such as a document without increasing computation complexity.
Existing approachs~\cite{chen2017reading,clark2017simple} usually first retrieve relevant passages with a ranking model and then return an answer with a reading model.
As for robustness, \newcite{wang2018robust} train the model with an adversarial data augmentation method.
\newcite{min2018efficient} propose to selectively read salient sentences rather than the entire passage, so as to avoid looking at the adversarial sentence.
Our approach, however, focuses on improving efficiency and robustness by transferring knowledge from a cumbersome ensemble model to a single model.

\noindent{\bf Knowledge Distillation.} 
Knowledge distillation is first explored by~\newcite{Bucilua06} and~\newcite{Hinton14}, which attempts to transfer knowledge defined as soft output distributions from a teacher to a student. 
Later works have been proposed to distill not only the final output but also intermediate representation from the teacher~\cite{Romero15,Zagoruyko17,Huang17}. 
\newcite{Papernot16} show that knowledge distillation can be used to prevent the network from adversarial attacks in image recognition.
\newcite{Radosavovic2017} introduce data distillation that annotates large-scale unlabelled data for omni-supervised learning.

In natural language processing (NLP), \newcite{Mou16} distill task-specific knowledge from word embeddings. 
\newcite{Kuncoro16} propose to learn a single parser from an ensemble of parsers.
\newcite{Kim16} investigate knowledge distillation for neural machine translation by approximately matching the sequence-level distribution of the teacher.
\newcite{Nakashole17} propose to learn bilingual mapping functions through a distilled training objective.
\newcite{Xu17} distill discriminative knowledge across languages for cross-lingual text classification.
Our work shows that the standard knowledge distillation and its novel variants can be successfully applied to the MRC task.
\section{Conclusion}
In this paper, we investigate knowledge distillation for machine reading comprehension.
We first explore vanilla knowledge distillation to transfer knowledge of answer positions, and then propose two variant approaches including answer distillation for penalizing student's predictions on confusing answer spans, and attention distillation for transferring teacher's attentive information.
Experiments show that the ensemble model has been successfully compressed into a single model that possesses better efficiency and robustness. 

In future work, we will explore new distillation methods that have better compression capabilities for MRC tasks, such as distilling knowledge from a single model instead of the ensemble without lossing performances, adding weights on knowledge based on the distilling quality and so on.
We also plan to further study the biased distillation problem and explore the compatibility of our approach in other NLP tasks such as natural language inference~\cite{Bowman15}, answer sentence selection~\cite{Yang15} and so on.

\section*{Acknowledgments}
We would like to thank Pranav Rajpurkar for his help with SQuAD submissions.
This work is funded by National Key R\&D Program of China (No. 0708063216003).

\bibliography{sections/reference}

\begin{thebibliography}{38}
\expandafter\ifx\csname natexlab\endcsname\relax\def\natexlab#1{#1}\fi

\bibitem[{Bahdanau et~al.(2015)Bahdanau, Cho, and Bengio}]{Bahdanau15}
Dzmitry Bahdanau, Kyunghyun Cho, and Yoshua Bengio. 2015.
\newblock Neural machine translation by jointly learning to align and
  translate.
\newblock In \emph{Proceedings of ICLR}.

\bibitem[{Bai et~al.(2010)Bai, Weston, Grangier, Collobert, Sadamasa, Qi,
  Chapelle, and Weinberger}]{bai2010learning}
Bing Bai, Jason Weston, David Grangier, Ronan Collobert, Kunihiko Sadamasa,
  Yanjun Qi, Olivier Chapelle, and Kilian Weinberger. 2010.
\newblock Learning to rank with (a lot of) word features.
\newblock \emph{Information retrieval}, 13(3):291--314.

\bibitem[{Bowman et~al.(2015)Bowman, Angeli, Potts, and Manning}]{Bowman15}
Samuel~R Bowman, Gabor Angeli, Christopher Potts, and Christopher~D Manning.
  2015.
\newblock A large annotated corpus for learning natural language inference
  systems.
\newblock In \emph{Proceedings of EMNLP}.

\bibitem[{Buciluǎ et~al.(2006)Buciluǎ, Caruana, and
  Niculescu-Mizil}]{Bucilua06}
Cristian Buciluǎ, Rich Caruana, and Alexandru Niculescu-Mizil. 2006.
\newblock Model compression.
\newblock In \emph{Proceedings of SIGKDD}, pages 535--541. ACM.

\bibitem[{Chen et~al.(2017)Chen, Fisch, Weston, and Bordes}]{chen2017reading}
Danqi Chen, Adam Fisch, Jason Weston, and Antoine Bordes. 2017.
\newblock Reading wikipedia to answer open-domain questions.
\newblock In \emph{Proceedings of ACL}.

\bibitem[{Clark and Gardner(2018)}]{clark2017simple}
Christopher Clark and Matt Gardner. 2018.
\newblock Simple and effective multi-paragraph reading comprehension.
\newblock In \emph{Proceedings of ACL}.

\bibitem[{Hermann et~al.(2015)Hermann, Kocisky, Grefenstette, Espeholt, Kay,
  Suleyman, and Blunsom}]{Hermann15}
Karl~Moritz Hermann, Tomas Kocisky, Edward Grefenstette, Lasse Espeholt, Will
  Kay, Mustafa Suleyman, and Phil Blunsom. 2015.
\newblock Teaching machines to read and comprehend.
\newblock In \emph{Proceedings of NIPS}.

\bibitem[{Hill et~al.(2016)Hill, Bordes, Chopra, and Weston}]{Hill16}
Felix Hill, Antoine Bordes, Sumit Chopra, and Jason Weston. 2016.
\newblock The goldilocks principle: Reading children’s books with explicit
  memory representations.
\newblock In \emph{Proceedings of ICLR}.

\bibitem[{Hinton et~al.(2014)Hinton, Vinyals, and Dean}]{Hinton14}
Geoffrey Hinton, Oriol Vinyals, and Jeff Dean. 2014.
\newblock Distilling the knowledge in a neural network.
\newblock In \emph{Proceedings of NIPS Workshop}.

\bibitem[{Hu et~al.(2018)Hu, Peng, Huang, Qiu, Wei, and Zhou}]{Hu17}
Minghao Hu, Yuxing Peng, Zhen Huang, Xipeng Qiu, Furu Wei, and Ming Zhou. 2018.
\newblock Reinforced mnemonic reader for machine reading comprehension.
\newblock In \emph{Proceedings of IJCAI}.

\bibitem[{Huang et~al.(2018)Huang, Zhu, Shen, and Chen}]{Huang17b}
Hsin-Yuan Huang, Chenguang Zhu, Yelong Shen, and Weizhu Chen. 2018.
\newblock Fusionnet: Fusing via fully-aware attention with application to
  machine comprehension.
\newblock In \emph{Proceedings of ICLR}.

\bibitem[{Huang and Wang(2017)}]{Huang17}
Zehao Huang and Naiyan Wang. 2017.
\newblock Like what you like: Knowledge distill via neuron selectivity
  transfer.
\newblock \emph{arXiv preprint arXiv:1707.01219}.

\bibitem[{Jia and Liang(2017)}]{Jia17}
Robin Jia and Percy Liang. 2017.
\newblock Adversarial examples for evaluating reading comprehension systems.
\newblock In \emph{Proceedings of EMNLP}.

\bibitem[{Joshi et~al.(2017)Joshi, Choi, Weld, and
  Zettlemoyer}]{joshi2017triviaqa}
Mandar Joshi, Eunsol Choi, Daniel~S Weld, and Luke Zettlemoyer. 2017.
\newblock Triviaqa: A large scale distantly supervised challenge dataset for
  reading comprehension.
\newblock In \emph{Proceedings of ACL}.

\bibitem[{Kadlec et~al.(2016)Kadlec, Schmid, Bajgar, and
  Kleindienst}]{Kadlec2016}
Rudolf Kadlec, Martin Schmid, Ondrej Bajgar, and Jan Kleindienst. 2016.
\newblock Text understanding with the attention sum reader network.
\newblock In \emph{Proceedings of ACL}.

\bibitem[{Kim and Rush(2016)}]{Kim16}
Yoon Kim and Alexander~M Rush. 2016.
\newblock Sequence-level knowledge distillation.
\newblock In \emph{Proceedings of EMNLP}.

\bibitem[{Ko{\v{c}}isk{\`y} et~al.(2017)Ko{\v{c}}isk{\`y}, Schwarz, Blunsom,
  Dyer, Hermann, Melis, and Grefenstette}]{Kovcisky2017}
Tom{\'a}{\v{s}} Ko{\v{c}}isk{\`y}, Jonathan Schwarz, Phil Blunsom, Chris Dyer,
  Karl~Moritz Hermann, G{\'a}bor Melis, and Edward Grefenstette. 2017.
\newblock The narrativeqa reading comprehension challenge.
\newblock \emph{arXiv preprint arXiv:1712.07040}.

\bibitem[{Kuncoro et~al.(2016)Kuncoro, Ballesteros, Kong, Dyer, and
  Smith}]{Kuncoro16}
Adhiguna Kuncoro, Miguel Ballesteros, Lingpeng Kong, Chris Dyer, and Noah~A
  Smith. 2016.
\newblock Distilling an ensemble of greedy dependency parsers into one mst
  parser.
\newblock In \emph{Proceedings of EMNLP}.

\bibitem[{Min et~al.(2018)Min, Zhong, Socher, and Xiong}]{min2018efficient}
Sewon Min, Victor Zhong, Richard Socher, and Caiming Xiong. 2018.
\newblock Efficient and robust question answering from minimal context over
  documents.

\bibitem[{Mou et~al.(2016)Mou, Jia, Xu, Li, Zhang, and Jin}]{Mou16}
Lili Mou, Ran Jia, Yan Xu, Ge~Li, Lu~Zhang, and Zhi Jin. 2016.
\newblock Distilling word embeddings: An encoding approach.
\newblock In \emph{Proceedings of CIKM}, pages 1977--1980. ACM.

\bibitem[{Nakashole and Flauger(2017)}]{Nakashole17}
Ndapandula Nakashole and Raphael Flauger. 2017.
\newblock Knowledge distillation for bilingual dictionary induction.
\newblock In \emph{Proceedings of EMNLP}, pages 2487--2496.

\bibitem[{Papernot et~al.(2016)Papernot, McDaniel, Wu, Jha, and
  Swami}]{Papernot16}
Nicolas Papernot, Patrick McDaniel, Xi~Wu, Somesh Jha, and Ananthram Swami.
  2016.
\newblock Distillation as a defense to adversarial perturbations against deep
  neural networks.
\newblock In \emph{2016 IEEE Symposium on Security and Privacy (SP)}, pages
  582--597. IEEE.

\bibitem[{Peters et~al.(2018)Peters, Neumann, Iyyer, Gardner, Clark, Lee, and
  Zettlemoyer}]{Elmo18}
Matthew~E. Peters, Mark Neumann, Mohit Iyyer, Matt Gardner, Christopher Clark,
  Kenton Lee, and Luke Zettlemoyer. 2018.
\newblock Deep contextualized word prepresentations.
\newblock In \emph{Proceedings of NACCL}.

\bibitem[{Radosavovic et~al.(2017)Radosavovic, Doll{\'a}r, Girshick, Gkioxari,
  and He}]{Radosavovic2017}
Ilija Radosavovic, Piotr Doll{\'a}r, Ross Girshick, Georgia Gkioxari, and
  Kaiming He. 2017.
\newblock Data distillation: Towards omni-supervised learning.
\newblock \emph{arXiv preprint arXiv:1712.04440}.

\bibitem[{Rajpurkar et~al.(2016)Rajpurkar, Zhang, Lopyrev, and
  Liang}]{Rajpurkar16}
Pranav Rajpurkar, Jian Zhang, Konstantin Lopyrev, and Percy Liang. 2016.
\newblock Squad: 100,000+ questions for machine comprehension of text.
\newblock In \emph{Proceedings of EMNLP}.

\bibitem[{Romero et~al.(2015)Romero, Ballas, Kahou, Chassang, Gatta, and
  Bengio}]{Romero15}
Adriana Romero, Nicolas Ballas, Samira~Ebrahimi Kahou, Antoine Chassang, Carlo
  Gatta, and Yoshua Bengio. 2015.
\newblock Fitnets: Hints for thin deep nets.
\newblock In \emph{Proceedings of ICLR}.

\bibitem[{Seo et~al.(2017)Seo, Kembhavi, Farhadi, and Hajishirzi}]{Seo17}
Minjoon Seo, Aniruddha Kembhavi, Ali Farhadi, and Hananneh Hajishirzi. 2017.
\newblock Bidirectional attention flow for machine comprehension.
\newblock In \emph{Proceedings of ICLR}.

\bibitem[{Tay et~al.(2018)Tay, Tuan, and Hui}]{Tay2018}
Yi~Tay, Luu~Anh Tuan, and Siu~Cheung Hui. 2018.
\newblock Multi-range reasoning for machine comprehension.
\newblock \emph{arXiv preprint arXiv:1803.09074}.

\bibitem[{Vinyals et~al.(2015)Vinyals, Fortunato, and Jaitly}]{Vinyals15}
Oriol Vinyals, Meire Fortunato, and Navdeep Jaitly. 2015.
\newblock Pointer networks.
\newblock In \emph{Proceedings of NIPS}.

\bibitem[{Wang and Jiang(2017)}]{Wang17a}
Shuohang Wang and Jing Jiang. 2017.
\newblock Machine comprehension using match-lstm and answer pointer.
\newblock In \emph{Proceedings of ICLR}.

\bibitem[{Wang et~al.(2018)Wang, Yan, and Wu}]{wang2018multi}
Wei Wang, Ming Yan, and Chen Wu. 2018.
\newblock Multi-granularity hierarchical attention fusion networks for reading
  comprehension and question answering.
\newblock In \emph{Proceedings of ACL}.

\bibitem[{Wang et~al.(2017)Wang, Yang, Wei, Chang, and Zhou}]{Wang17b}
Wenhui Wang, Nan Yang, Furu Wei, Baobao Chang, and Ming Zhou. 2017.
\newblock Gated self-matching networks for reading comprehension and question
  answering.
\newblock In \emph{Proceedings of ACL}.

\bibitem[{Wang and Bansal(2018)}]{wang2018robust}
Yicheng Wang and Mohit Bansal. 2018.
\newblock Robust machine comprehension models via adversarial training.
\newblock In \emph{Proceedings of NAACL}.

\bibitem[{Xiong et~al.(2018)Xiong, Zhong, and Socher}]{Xiong17}
Caiming Xiong, Victor Zhong, and Richard Socher. 2018.
\newblock Dcn+: Mixed objective and deep residual coattention for question
  answering.
\newblock In \emph{Proceedings of ICLR}.

\bibitem[{Xu and Yang(2017)}]{Xu17}
Ruochen Xu and Yiming Yang. 2017.
\newblock Cross-lingual distillation for text classification.
\newblock In \emph{Proceedings of ACL}.

\bibitem[{Yang et~al.(2015)Yang, Yih, and Meek}]{Yang15}
Yi~Yang, Wen-tau Yih, and Christopher Meek. 2015.
\newblock Wikiqa: A challenge dataset for open-domain question answering.
\newblock In \emph{Proceedings of EMNLP}.

\bibitem[{Yu et~al.(2018)Yu, Dohan, Luong, Zhao, Chen, Norouzi, and Le}]{Yu18}
Adams~Wei Yu, David Dohan, Minh-Thang Luong, Rui Zhao, Kai Chen, Mohammad
  Norouzi, and Quoc~V Le. 2018.
\newblock Qanet: Combining local convolution with global self-attention for
  reading comprehension.
\newblock In \emph{Proceedings of ICLR}.

\bibitem[{Zagoruyko and Komodakis(2017)}]{Zagoruyko17}
Sergey Zagoruyko and Nikos Komodakis. 2017.
\newblock Paying more attention to attention: Improving the performance of
  convolutional neural networks via attention transfer.
\newblock In \emph{Proceedings of ICLR}.

\end{thebibliography}
\bibliographystyle{acl_natbib_nourl}

\end{document}